\documentclass[10pt,twocolumn,letterpaper]{article}

\usepackage{btas}
\usepackage{times}
\usepackage{epsfig}
\usepackage{graphicx}
\usepackage{amsmath}
\usepackage{amssymb}
\usepackage{booktabs, multicol, multirow}
\usepackage{array}
\usepackage{color}
\usepackage{subfigure}
\usepackage{caption}
\usepackage{cite}
\usepackage[norule,symbol,perpage]{footmisc}

\btasfinalcopy 
\newcommand{\comment}[1]{\textcolor{black}{#1}}

\ifbtasfinal\fi
\makeatletter  
\def\ps@IEEEtitlepagestyle{  
\def\@oddfoot{\mycopyrightnotice}  
\def\@evenfoot{}  
}  
\def\mycopyrightnotice{  
{\hfill \footnotesize 978-1-7281-1522-1/19/\$31.00 \copyright 2019 IEEE\hfill}  
}  
\makeatother
\begin{document}

\title{Zero-Shot Deep Hashing and Neural Network Based Error Correction \\ for Face Template Protection}

\author{Veeru Talreja, \ Matthew C. Valenti, and \ Nasser M. Nasrabadi\\
West Virginia University\\
Morgantown, WV, USA\\
{\tt\small vtalreja@mix.wvu.edu,valenti@ieee.org,nasser.nasrabadi@mail.wvu.edu }
}

\maketitle
\thispagestyle{empty}

\begin{abstract}
   In this paper, we present a novel architecture that integrates a deep hashing framework with a neural network decoder (NND) for application to face template protection. It improves upon existing face template protection techniques to provide better matching performance with one-shot and multi-shot enrollment. A key novelty of our proposed architecture is that the framework can also be used with zero-shot enrollment. This implies that our architecture does not need to be re-trained even if a new subject is to be enrolled into the system. The proposed architecture consists of two major components: a deep hashing (DH) component, which is used for robust mapping of face images to their corresponding intermediate binary codes, and a NND component, which corrects errors in the intermediate binary codes that are caused by differences in the enrollment and probe biometrics due to factors such as variation in pose, illumination, and other factors. The final binary code generated by the NND is then cryptographically hashed and stored as a secure face template in the database.  The efficacy of our approach with zero-shot, one-shot, and multi-shot enrollments is shown for CMU-PIE, Extended Yale B, WVU multimodal and Multi-PIE face databases. With zero-shot enrollment, the system achieves approximately 85\% genuine accept rates (GAR) at $0.01\%$ false accept rate (FAR), and with one-shot and multi-shot enrollments, it achieves approximately 99.95\% GAR at $0.01\%$ FAR, while providing a high level of template security.        
\end{abstract}
\let\thefootnote\relax\footnotetext{\mycopyrightnotice}
\vspace{-0.25cm}

\section{Introduction}
\vspace{-0.15cm}


The leakage of biometric information, such as a stored template, to an adversary constitutes a serious threat to security and privacy because if an adversary gains access to a biometric database, he can potentially obtain the stored user information \cite{talreja_2018_using,talreja_icce_2018,sobhan_btas}. The attacker can use this information to gain unauthorized access to a system, abuse the biometric information for unintended purposes, and violate user privacy \cite{nagar_multibiometriccryptosystems_2012}. Hence, biometric template protection is an important issue and the main focus of this paper.

 For a template to be secure, it must satisfy the important properties of \emph{noninvertibility} and \emph{cancelability}. Noninvertibility implies that it must be computationally difficult to recover the original biometric data when a template is given (e.g., compromised). Cancelability implies that if a template gets compromised, it should be possible to revoke the compromised template and generate a new template using a different transformation. The fundamental challenge in designing a biometric template protection scheme satisfying the above properties is the high intra-user variability that occurs due to signal variations in the multiple acquisitions of the same biometric trait and also low inter-user variability \cite{talreja_globalsip_2017}. The high intra-user variability leads to high false reject rate (FRR), while low inter-user variability leads to high false accept rate (FAR) \cite{talreja_icc_2019,sobhan_icb}. 

Prior work in face template protection has tried to decrease the intra-user variability and increase the inter-user variability by using multiple acquisitions of the user's biometric trait (multi-shot) during enrollment \cite{pandey_2015_face, pandey_deep_2016, Jindal_2017_face}.
 Pandey and Govindaraju \cite{pandey_2015_face} extract features from selected local regions of the face followed by quantization and cryptographic hashing to generate a final template. Although the scheme benefits from secure hash functions, it suffers from low matching accuracy. To overcome the shortcomings in \cite{pandey_2015_face}, Pandey \etal \cite{pandey_deep_2016} provide another face template protection algorithm, where unique maximum entropy binary (MEB) codes are assigned to each user and these MEB codes are used as labels to train a convolutional neural network (CNN) and learn the mapping from face images to MEB codes. The MEB code assigned to each user is cryptographically hashed and stored as a template in the database. This algorithm \cite{pandey_deep_2016} suffers from a high FRR for higher matching accuracy and moreover, it is only compatible with multi-shot enrollment. To improve upon this algorithm, Jindal \etal \cite{Jindal_2017_face}  use a deeper and better CNN for robust mapping of face images to binary codes with significantly better matching performance and compatibility with both one-shot and multi-shot enrollment.

 In \emph{one-shot} enrollment, strictly only one image is used during enrollment and training, while in \emph{multi-shot} enrollment, multiple images are used during enrollment and training of the network. However, both the deep learning based methods \cite{pandey_deep_2016, Jindal_2017_face} are not compatible with \emph{zero-shot} enrollment, wherein a subject not seen during training needs to be enrolled into the system.  For both of the above cited methods,  whenever a new subject needs to be enrolled, the complete network needs to be retrained with the new subject included into the training database.



To address the above problems, we propose an architecture for face template protection by integrating a deep hashing (DH) framework with a neural network decoder (NND). Deep hashing is the application of deep learning to generate compact binary vectors from raw image data and is generally used for fast image retrieval  \cite{taherkhani_2018_deep, taherkhani_2018_facial, learning_lin_2016,deep_hashing_liu_2016,cao_2017_hashnet,zhu_2016_deep,Yuan_2018_ECCV,Yuan_2018_DeepHV,kazemi_nips_2018,kazemi_biosig_2018}. In addition to using deep hashing to generate binary codes, we use error correcting codes (ECC) as an additional component. ECC is used to compensate for the difference in enrollment and probe biometrics (arising from variation in pose, illumination, noise in biometric capture). Recent work has shown that the same kinds of neural network architectures used for classification can also be used to decode ECC codes \cite{2016_nachmani_NND,Lugosch_2017_NeuralOM,nachmani2018deep,gruber2017deep}. In this work, we integrate a NND \cite{2016_nachmani_NND} into our deep hashing architecture as an ECC component to improve the matching performance. 

Specifically, our proposed architecture consists of two major components: a DH component, which is used for robust mapping of face images to their intermediate binary codes, and a NND component, which corrects errors in the intermediate binary codes that are caused by differences in the enrollment and probe biometrics due to factors such as variation in pose, illumination, and other factors. The final binary code generated by the NND component is then cryptographically hashed and stored as the  secure  face  template  in  the  database. We have used SHA3-512 as the cryptographic hash function, since it is a current standard for string-based passwords and provides strong security. The template generated after cryptographic hashing has no correlation with the binary codes generated at the output of the NND. To improve the template security, we have also optimized our deep hashing architecture by using an additional loss function to maximize the entropy of the binary codes being generated, which also helps to minimize the intra-user variability and maximize the inter-user variability.     

The advantage of using a NND instead of a conventional ECC decoder is that implementing the decoder as  a  neural  network  provides the benefit of using a similar architecture as the DH component that generates the binary code,  and  hence,  it  can  be  more  efficiently  jointly optimized  and  implemented  within  a  common  framework. Another advantage of using the NND is that it provides an opportunity to jointly learn and optimize with respect to biometric datasets, which  are  not  necessarily  characterized by  Gaussian noise as is assumed by a conventional decoder. Motivated by this, in this paper, we have integrated our DH network with the NND by using a joint optimization process to formulate our novel face template protection architecture. This proposed architecture  can also be used with zero-shot enrollment while still offering the potential to improve the matching performance with one-shot and multi-shot enrollment. 

To summarize, the main contributions of this paper are:

1. The development of a facial authentication system that uses protected templates and can be used for zero-shot enrollment where a new subject not present during the training can be enrolled into the system without a need to retrain the deep learning framework.  

2. Inclusion and optimization of the neural network based decoder to compensate for the distortion in biometric measurements and the end-to-end joint optimization of the overall system.

3. Overall improvement in the architecture for face template protection provided by the state-of-the-art \cite{Jindal_2017_face,pandey_deep_2016} to achieve comparable matching performance with one-shot and multi-shot enrollment on PIE, Extended Yale B and Multi-PIE databases.
    
    \vspace{-0.25cm}

\section {Proposed Architecture} \label{sec:arch}
\begin{figure}[t]
\vspace{-0.30cm}
\centering
\includegraphics[width=8.25cm]{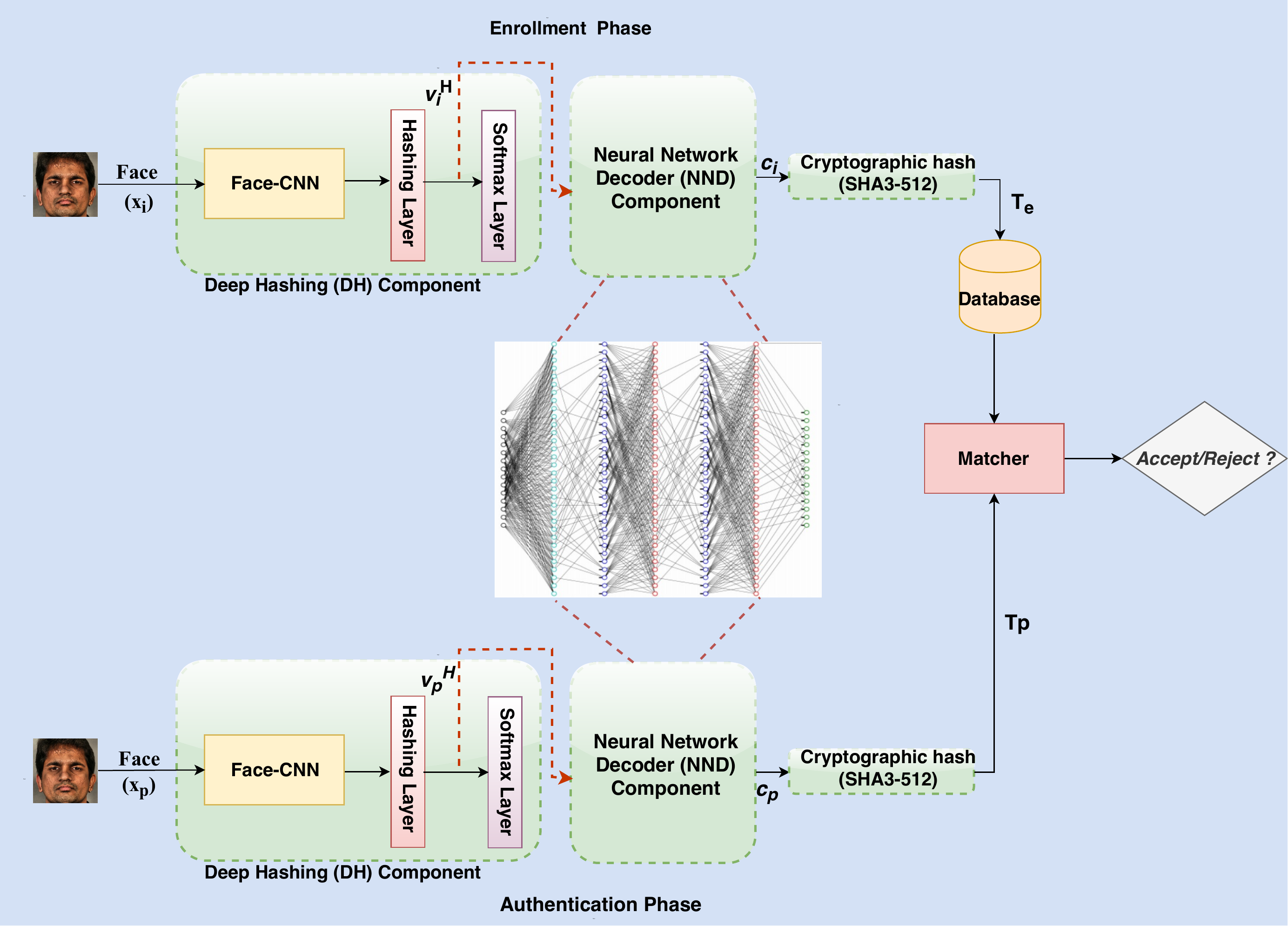}
\vspace{-0.30cm}
\caption{Block diagram of the proposed system. The NND is shown in the figure. The DH is shown in Fig. \ref{fig:arch_bla}}\label{fig:enrol}
\vspace{-0.45cm}

\end{figure}

\vspace{-0.15cm}
In this section, we present a system overview of the proposed architecture, which is shown in Fig. \ref{fig:enrol}, and also present the enrollment and authentication procedure. The architecture consists of two important components: a deep hashing component and a neural network decoder component. In this paper, we have implemented this architecture for face biometrics. However this system could also be extended for use with other biometrics such as iris or fingerprint or a combination of multiple biometrics.

\subsection{Deep Hashing Component} \label{subsec:dh}

The main function of the deep hashing (DH) component (which can interchangeably be called a deep hashing network) is to map the input facial images to binary codes. These binary codes are not pre-defined as in \cite{Jindal_2017_face,pandey_deep_2016} but rather are generated as an output of the DH component. This is one of the reasons why this framework can be used with zero-shot enrollment as well. Let $\textbf{X} = \{\textbf{x}_i\}_{i=1}^{N}$ denote $N$ facial images and $\textbf{Y} = \{\textbf{y}\textsubscript{i}\in \{0,1\}^M\}_{i=1}^{N}$
be their associated label vectors, where $M$ denotes the number of class labels. An entry of the vector \textbf{y}\textsubscript{i}  is
1 if an image \textbf{x}\textsubscript{i} belongs to the corresponding class and
0 otherwise. The goal of the DH component is to learn a mapping or a hash function $\mathcal{G}:\textbf{X} \longrightarrow \{0,1\}^{K\times N}$, which maps a set of images to their $K$-bit binary codes $\textbf{C}=\{\textbf{c}\textsubscript{i}\} \in \{0,1\}^{K\times N}$, while preserving the semantic
similarity among image data. Specifically, in the DH component, we deploy a supervised hashing algorithm that exploits semantic labels to create binary codes with the following required properties:
(1) The semantic similarity between image labels is preserved in the binary codes; images that share common class labels are mapped to same (or close) binary codes. (2) The bits in a code are evenly distributed, which means that the value of each bit is equally likely to be 0 or 1, leading to high entropy and discriminative binary codes.

With recent advances in deep learning, hash functions can be constructed using a CNN that is capable of learning semantic representations from input images. Our approach is built on  the existing deep VGG-19 model \cite{simonyan_very_deep_2014}. The advantage of our approach is that it can be implemented using other deep models as well, such as AlexNet \cite{krizhevsky_imagenet_2012}. We introduce our approach based on VGG-19. For our architecture, we use only the first 16 convolutional layers of pre-trained VGG-19. To the 16 convolutional layers, we add our own fully connected layer \emph{fc1} and an output softmax layer. The length of the \emph{fc1} layer depends upon the size of the binary code $K$  that we want to use for template generation. These 16 convolutional layers and the 1 fully connected layer \emph{fc1} will be termed as ``Face-CNN".   



\begin{figure}[t]
\vspace{-0.30cm}
\centering
\includegraphics[width=8.5cm]{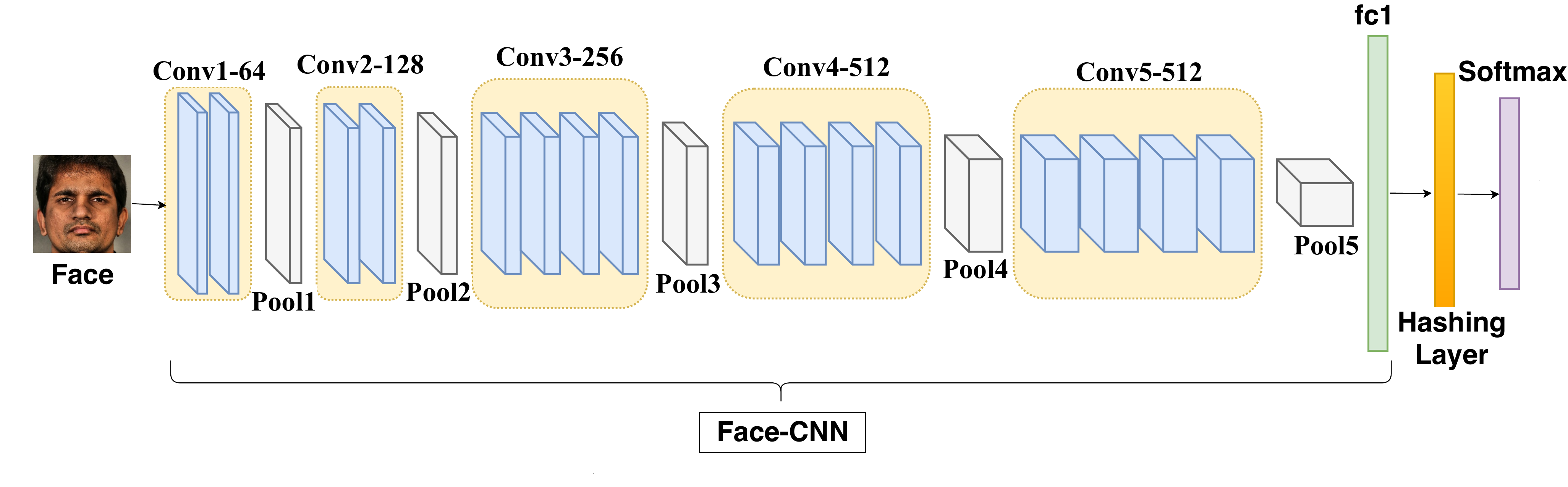}
\vspace{-0.75cm}
\caption{Proposed Deep Hashing network.}\label{fig:arch_bla}
\vspace{-0.45cm}
\end{figure}

The output of the Face-CNN at \emph{fc1} is a feature vector of unquantized values. This output at \emph{fc1} can be directly binarized by thresholding at any numerical value or thresholding at the population mean. However, this kind of thresholding leads to a quantization loss, which results in sub-optimal binary codes. To account for this quantization loss and incorporate the deep representations into the hash function learning, we add a latent layer called hashing layer $H$ with $K$ units to the top of layer \emph{fc1} (i.e., the layer immediately before the output layer), as illustrated in Fig. \ref{fig:arch_bla}. This hashing layer is fully connected to \emph{fc1} and uses the sigmoid activation function so that the outputs are between 0 and 1. The main purpose of the hashing layer is to capture the quantization loss incurred while converting the extracted face features (output of \emph{fc1} layer) into binary codes. 

Let $\textbf{W}^{(H)} \in \mathbb{R} ^{d\times K} $ denote the weights between \emph{fc1} and the latent layer. For a given image $\textbf{x}_i$ with the feature vector $\textbf{v}_i^{(fc1)} \in \mathbb{R}^{d} $ in layer \emph{fc1}, the activations of the units in H is given as $\textbf{v}_{i}^{(H)}=\sigma(\textbf{v}_i^{(fc1)}\textbf{W}^{(H)}+b^{(H)})$, where $\textbf{v}_{i}^{(H)}$ is a $K$-dimensional vector, $b^{(H)}$ is the bias term and $\sigma(.)$ is the sigmoid activation defined as $\sigma(z)=1/(1+\exp (-z))$, with $z$ a real value. 

The combination of the Face-CNN, the hashing layer and the output softmax layer forms the DH network. The details about the training of the DH network are given in Sec. \ref{subsec:stage1}

\subsection{Neural Network Decoder Component}

After training the DH network, we can directly binarize the output of the hashing layer using a threshold of $0.5$ and use the result as a binary code. Henceforth, we will refer to the output of the hashing layer from the DH component as an \emph{intermediate binary code}. At this stage, we can cryptographically hash the intermediate binary code and store it as a secure template. However, cryptographic hashes are extremely sensitive to noise and there is always some inherent noise and distortion in biometric measurements such as variations in pose, illumination, or noise due to the biometric capturing device, leading to differences in enrollment and probe biometrics from the same subject. Due to these differences, enrollment and probe biometrics from the same subject may lead to different intermediate binary codes at the output of the hashing layer, which may result in the cryptographic hash mismatch and excessive false rejections. We need to compensate for this distortion in biometric measurements to make the system more robust, improve the matching performance, and provide another layer of security.  This is achieved by using error-correcting codes (ECC). ECC can compensate for the biometric distortion by forcing the enrollment and probe biometric to decode to the same message, making the system more robust to noise in the biometric measurements, which helps in improving the matching performance.

\comment{While we could use a conventional ECC decoder at the output of the hashing layer, such a decoder is only suitable when the codewords are corrupted by Gaussian noise. Generally, the distortion in biometric measurements may not necessarily be characterized by Gaussian noise. For this reason, we need to be able to train our ECC decoder to be optimized for biometric measurements.}

Recent research in the field of ECC has focused on designing a neural network architecture as an ECC decoder\cite{2016_nachmani_NND,Lugosch_2017_NeuralOM}. We can adapt such a neural network based ECC decoder, train it, and use it as an ancillary component to refine the intermediate binary codes generated by hashing layer in the DH component. The advantage of using NND instead of a conventional decoder is that it allows for a common architectural framework to be used for both the hashing framework (i.e., the DH) and the decoder, and it provides an opportunity to jointly learn and optimize with respect to biometric datasets, which are not necessarily characterized simply by Gaussian noise as is assumed by conventional decoders. For our application, we have chosen the NND described in \cite{2016_nachmani_NND} as the basis for our neural network ECC decoder to be integrated with the DH component. Due to space limitations, we do not provide full details on the operation of NND, as details can be found in the original paper. Rather, we focus our discussion on how the NND decoder is integrated with our architecture and also discuss the differences in training the NND relative to \cite{2016_nachmani_NND}.

\subsection{Enrollment and Authentication} \label{subsec:enroll}

 During enrollment, the facial image of the user is captured, resized and is given as input ($\textbf{x}\textsubscript{i}$) to the trained DH network as shown in Fig. \ref{fig:enrol}. The intermediate binary code $\textbf{v}\textsubscript{i}^\textbf{(H)}$ for the user is generated at the output of the hashing layer of the DH network. This intermediate binary code is fed through a trained NND network and the final binary code $\textbf{c}\textsubscript{i}$ for a given user is generated by simple thresholding of the output of NND at 0.5. The final binary code is cryptographically hashed using SHA3-512 to create the enrollment face template $\textbf{T}\textsubscript{e}$ to be stored in the database. The final binary code is not provided to the user or stored in an unprotected form. Only the cryptographic hash of the final binary code is stored as a template in the database. During the authentication phase, a new sample of the enrolled user is fed through the DH network and the output of the hashing layer is fed through the NND to get the final binary code $\textbf{c}\textsubscript{p}$  for the probe. This final binary code is cryptographically hashed using SHA3-512 to generate the probe template $\textbf{T}\textsubscript{p}$, which is compared with the enrollment template $\textbf{T}\textsubscript{e}$ in the matcher to generate a binary score of accept/reject nature.

\vspace{-0.25cm}
\section{Training of the Proposed Architecture}
The architecture described in Sec. \ref{sec:arch} is trained in three stages. In Stage 1, we use a novel loss function to train and learn the parameters of the DH component to generate intermediate binary codes at the output of the hashing layer; in Stage 2, the intermediate binary codes from Stage 1 are passed through a conventional ECC decoder to generate the ground truth, which will be used to fine-tune a neural network decoder (NND); in Stage 3, the NND decoder is trained using the ground truth from Stage 2 and this NND is then integrated with the DH component followed by a joint optimization of the overall system. 

\vspace{-0.30cm}
\subsection{Stage 1: Training the DH component} \label{subsec:stage1}

As discussed in the Sec. \ref{sec:arch}, the DH component consists of Face-CNN (\emph{Conv1-Conv5}+\emph{fc1}), hashing layer $H$ and the output softmax layer. Before adding the hashing layer to the DH component, the Face-CNN with the output softmax layer is trained with the CASIA-Webface \cite{yi_learning_2014}, which contains 494,414 facial images corresponding to 10,575 subjects. For training the Face-CNN, all the raw facial images are first aligned in 2-D and cropped to a size of $224\times 224$ before passing through the network \cite{dlib_09}. The only other pre-processing is subtracting the mean RGB value, computed on the training set, from each pixel. The training is carried out by optimizing the multinomial logistic regression objective using mini-batch gradient descent with momentum. The number of nodes in the fully connected layer \emph{fc1} before the softmax layer depends upon the size of the binary code $K$. If $K$ is 255, then the length of \emph{fc1} is 512, and if $K$ is 1023, then the length of \emph{fc1} is 2048. 


After the training of Face-CNN, the hashing layer is added on top of the \emph{fc1} layer (i.e. just before the output softmax layer) to form the DH component. The database used for training the DH component have been discussed in detail in Sec. 4. For training the full DH component, we have used a novel objective function, which helps in reducing the quantization loss and also maximize the entropy to generate optimal and discriminative binary codes at the output of the hashing layer. The objective function used for training the DH component is a combination of classification loss, quantization loss and entropy maximization loss. The classification loss has been added into the DH network by using the \emph{softmax} layer as shown in Fig. \ref{fig:arch_bla}. $E_{1}(\textbf{w})$ denotes the objective function required to fulfill the classification task: 
\vspace{-0.20cm}
\begin{equation} E_{1}(\textbf{w})=\frac{1}{N}\sum_{n=1}^{N}L_i(f(x_{i},\textbf{w}),y_{i}) + \lambda ||\textbf{w}||_{2}^{2} \label{eq:2},\vspace{-0.25cm}\end{equation} where the first term $L_i(.)$ is the classification loss for a training instance $i$ and is described below, $N$ is the number of training images in a mini-batch. $f(x_{i},\textbf{w})$ is the predicted softmax output of the network and is a function of the input training image $x_i$ and the weights of the network \textbf{w}. The second term is the regularization function where $\lambda$ governs the relative importance of the regularization. 
Let the predicted softmax output $f(x_{i},\textbf{w})$ be denoted by $\hat{y}_{i}$. The classification loss for the $i^\mathsf{th}$ training instance is given as: 
\vspace{-0.20cm}
\begin{equation}L_{i}(\hat{y}_{i},y_{i})=-\sum_{m=1}^{M}y_{i,m}\ln \hat{y}_{i,m} \label{eq:3},\vspace{-0.25cm}\end{equation} where $y_{i,m}$ and $\hat{y}_{i,m}$ is the ground truth and the prediction result for the $m^\mathsf{th}$ unit of the $i^\mathsf{th}$ training instance, respectively and $M$ is the number of output units.


The output of the hashing layer is a $K$-dimensional vector denoted by $\textbf{v}^{(H)}_{i}$, corresponding to the $i$-th input image. The $n$-th element of this vector is denoted by $v^{(H)}_{i,n} (n=1,2,3, \cdots,K)$. The value of $v^{(H)}_{i,n}$ is in the range of $[0,1]$ because it has been activated by the $\mbox{sigm}$ activation. To capture the quantization loss of thresholding at 0.5 and make the codes closer to either 0 or 1, we add a constraint of binarization loss between the hashing layer activations and 0.5, which is given by $\sum_{i=1}^{N}||\textbf{v}^{(H)}_{i}-0.5\textbf{e}||^{2}$, where $N$ is the number of training images in a mini-batch and \textbf{e} is the $K$-dimensional vector with all elements equal to 1.  Let $E_{2}(\textbf{w})$ denote this constraint to boost the activations of the units in the hashing layer to be closer to 0 or 1 and this constraint needs to be maximized in order to push the binary codes closer to 0 and 1:
\vspace{-0.45cm}
\begin{equation}E_{2}(\textbf{w})=-\frac{1}{K}\sum_{i=1}^{N}||\textbf{v}^{(H)}_{i}-0.5\textbf{e}||^{2} \label{eq:4}.\vspace{-0.25cm}\end{equation}  

In the state-of-the-art face template protection methods \cite{pandey_deep_2016,Jindal_2017_face}, pre-defined maximum entropy binary codes have been used as labels to train the deep CNNs. Maximum entropy an important requirement for improving the discrimination as well as improving the security. To include this requirement into our architecture, it is important that the binary codes at the output of the hashing layer have equal number of 0's and 1's, which maximizes the entropy of the discrete distribution and results in binary codes with better discrimination. Let $E_{3}(\textbf{w})$ denote the loss function that forces the output of each node to have a $50\%$ chance of being 0 or 1; $E_{3}(\textbf{w})$  needs to be minimized to maximize the entropy:
\vspace{-0.55cm}
\begin{equation}E_{3}(\textbf{w})= \sum_{i=1}^{N}(\text{mean}(\textbf{v}^{(H)}_{i})-0.5)^{2}\label{eq:5}.\vspace{-0.25cm}\end{equation}
The overall objective function to be minimized for a semantics-preserving efficient binary codes is given as:
\vspace{-0.15cm}
\begin{equation}\alpha E_{1}(\textbf{w})+ \beta E_{2}(\textbf{w}) + \gamma E_{3}(\textbf{w}) \label{eq:6},\end{equation} where $\alpha$, $\beta$, and $\gamma$ are the tuning parameters of each term. 



The objective function given in (\ref{eq:6}) can be minimized by using stochastic gradient descent (SGD) efficiently by dividing the training samples into batches.


\subsection{Stage 2: Generating the Ground Truth for Training the Neural Network Decoder}

 As already mentioned, we have used the NND from \cite{2016_nachmani_NND} as our ECC component for added security and also to make the system robust to variations in biometric measurements, which would make the architecture applicable even for zero-shot enrollment. The NND in \cite{2016_nachmani_NND} is optimized for a Gaussian noise channel and the database used for training the NND reflects various channel output realizations when the zero codeword has been transmitted. However, for our proposed system, the NND needs to be optimized for use with biometric data, where the channel noise is characterized by the image distortions (e.g., pose variations, illumination variations and noise due to biometric capturing device) in different biometric images of the same subject. We can create the input dataset for training the NND by using the different facial images for the same subject. However, there is a major issue that needs to be addressed for training the NND with the facial images, and that is we do not have the labels or the ground truth codewords that these input images need to be mapped to. An external conventional ECC decoder can be used to generate the ground truth codewords for this input dataset of facial images.
 
 

After training DH network in Stage 1, a number of facial images of subjects disjoint from the subjects used for training the DH network are used for generating the ground truth for training the NND. First, we use the trained DH network to extract the binary vectors at the output of the hashing layer (threshold the sigmoid activations at 0.5) for this disjoint dataset. These extracted binary vectors are used as input to a conventional ECC decoder for soft-decision decoding. Hard-limiting the output of the ECC decoder generates ground truth codewords that are used as labels for optimizing the NND in Stage 3. While usually all the binary vectors of a given subject are mapped by the decoder to the same codeword, it is possible that the vectors could be mapped to different codewords.  This is especially the case when there are substantial differences in the variations of the facial images for a given subject, or when the ECC code is not sufficiently strong. In the case that the input binary vectors get mapped to a plurality of codewords, the most common of these codewords is used as the ground truth for that subject. 

\vspace{-0.25cm}

\subsection{Stage 3: Joint Optimization of Deep Hashing and Neural Network Decoder}

In Stage 3 of the training, first we train the NND using the same procedure and database outlined in the original paper \cite{2016_nachmani_NND}. Next, the NND is fine-tuned for our facial biometric data. For fine-tuning the NND, we use the same disjoint dataset that was used for the ECC coventional decoder in Stage 2. Similar to Stage 2, the input to NND is given by the feature vectors generated at the output of the hashing layer of the DH network and the labels are provided by the decoded codewords generated by the conventional ECC decoder in Stage 2. We have used sigmoid activation for the last layer of NND so that the final network output is in the range $[0,1]$. This makes it possible to train and fine-tune the NND using binary cross-entropy loss function:
\vspace{-0.30cm}
\begin{equation}
    L(o,y)=-\frac{1}{N}\sum_{i=1}^{N} y_{i}\log(o_i)+(1-y_i)\log(1-o_i), \label{eq:7}\vspace{-0.25cm}
\end{equation} where $o_i$, $y_i$ are the actual $i$th  component  of  the NND  output and ground truth codeword (label), respectively. 

After fine-tuning the NND, we integrate DH and NND by discarding the softmax layer in the DH network and connecting the output of the hashing layer from DH as input to NND to create and end-to-end face template protection architecture. This overall system is then optimized end-to-end using the same dataset as used for fine-tuning NND and also the same cross-entropy loss function given in (\ref{eq:7}).

Indeed a key benefit of using NND over conventional ECC decoding is that the NND can be trained to force all the binary vectors of the same subject to be decoded to the corresponding common codeword. This also helps for zero-shot enrollment, where, even for the subjects not seen during training, the trained NND will generally compensate for the biometric distortion by forcing the enrollment and probe biometric of a subject to decode to a common codeword.



\section{Implementation and Evaluation}
\vspace{-0.15cm}
\subsection{Databases and Data Augmentation}\label{subsec:augm}
  We use the following databases for our training and testing of our proposed system. \comment{The first 3 datasets have been used for a fair comparison with the state-of-the-art in face template protection. The last dataset is used to test our system on a modern large scale face recognition dataset}:

1. The CMU PIE \cite{Sim_cmu-pie_2002} database consists of 41,368 images of 68 subjects. These images have been taken under 13 different poses, 43 different illumination conditions, and 4 different expressions. We use 5 poses (p05, p07, p09, p27, p29) and all illuminations variations for our training and testing. We have used 50 subjects for training the DH network with the hashing layer. Out of the remaining 18 subjects, 12 subjects have been used for training the NND and fine-tuning the overall system end-to-end, and the remaining 6 subjects have been used for testing of zero shot enrollment.  For one shot and multi-shot, the train and test data split is consistent with \cite{Jindal_2017_face} for the 50 subjects.

2. The extended Yale Face \cite{georghiodes_yale_2001} database contains 2,432 images corresponding to 38 subjects with frontal pose and under different illumination conditions. We have used the cropped version of the database as used in \cite{pandey_deep_2016}. Out of the 38 subjects, we have used 22 subjects for training the DH network with the hashing layer. Out of the remaining 16 subjects, 10 subjects have been used for training the NND and fine-tuning the overall system from end-to-end, and the remaining 6 subjects have been used for testing of zero shot enrollment. Again, for multi-shot enrollment, the train and test data split is consistent with \cite{pandey_deep_2016} for the 22 subjects. 

3. For the CMU Multi-PIE \cite{gross_multi-pie} database, we have just used it to test the robustness of our overall system to change in session and lighting conditions as in \cite{pandey_deep_2016}. CMU Multi-PIE database contains 750,000 images corresponding to 337 subjects under 4 different sessions, 15 view points and 19 illumination conditions. As in \cite{pandey_deep_2016}, we have used the session 3 and session 4 with total of 198 common subjects. We have chosen session 3 for enrollment and session 4 for testing. We have used 125 subjects for training the DH network with the hashing layer. Out of the remaining 73 subjects, 53 subjects have been used for training the NND and fine-tuning the overall system from end-to-end, and the remaining 20 subjects have been used for testing of zero shot enrollment. 

4. \comment{The WVU multimodal dataset \cite{wvu_multimodal_2017} was collected at West Virginia University in year 2012 and 2013. The data for the year 2013 and 2012 contain 61,300 and 70,100 facial images in different poses corresponding to 1063 and 1200 subjects, respectively. There are 294 common subjects in year 2012 and 2013 data. The unique 769 (1063-294) subjects from 2013 dataset are used for training the DH component. The remaining 294 common subjects from 2013 dataset are used for training the NND and fine-tuning the overall system from end-to-end. For one shot, we randomly select one image per user for training and the rest are used for testing. For multi-shot, 10 images are randomly chosen for training and the rest are used for testing. A subset of 50 subjects of the unique 900 subjects from the 2012 dataset are used for testing of zero shot enrollment.}

 In order to have sufficient data for training our deep learning algorithm, we use data augmentation. For each facial image, we apply horizontal flip, scaling to $60\%,70\%,80\%$, and $90\%$ to generate five augmented images \cite{osahor_2019_design,ferdous_2019_super}. For each augmented image of size $m\times m$ we extract all possible crops of size $n\times n$ yielding as total of $(m-n+1) \times (m-n+1)$ crops. Therefore, data augmentation yields a total of $5\times(m-n+1)\times (m-n+1)$ images for each face image. For our experiments, we have chosen value of $m=224$ and $n=221$.

\vspace{-0.30cm}

\subsection{Details of the Code and Decoder}
The intermediate binary code generated at the output of hashing layer of the DH network is considered to be the noisy codeword of some error correcting code (ECC) that  we  can  select and this noisy codeword can be decoded using the NND to generate the final hash code that is cryptographically hashed and stored as template in the database. We have used a BCH code as an ECC for our experiments. The size of the BCH code that we can use for NND depends upon the size of the intermediate binary code that we want. For comparison with the state-of-the-art methods in \cite{pandey_deep_2016,Jindal_2017_face}, we have used 255 and 1023 as the size of the intermediate binary codes, which is also the size of the final binary code used. The BCH codes that we have used for NND are BCH(255,187), and BCH(1023,933). For the external ECC decoder, we have used the BCH decoder from the communication toolbox in MATLAB\textsuperscript{\textregistered}.




\begin{figure*}[t]
\centering     
\subfigure[CMU-PIE ]{\label{fig:a}\includegraphics[width=6.5cm,height=2.7cm]{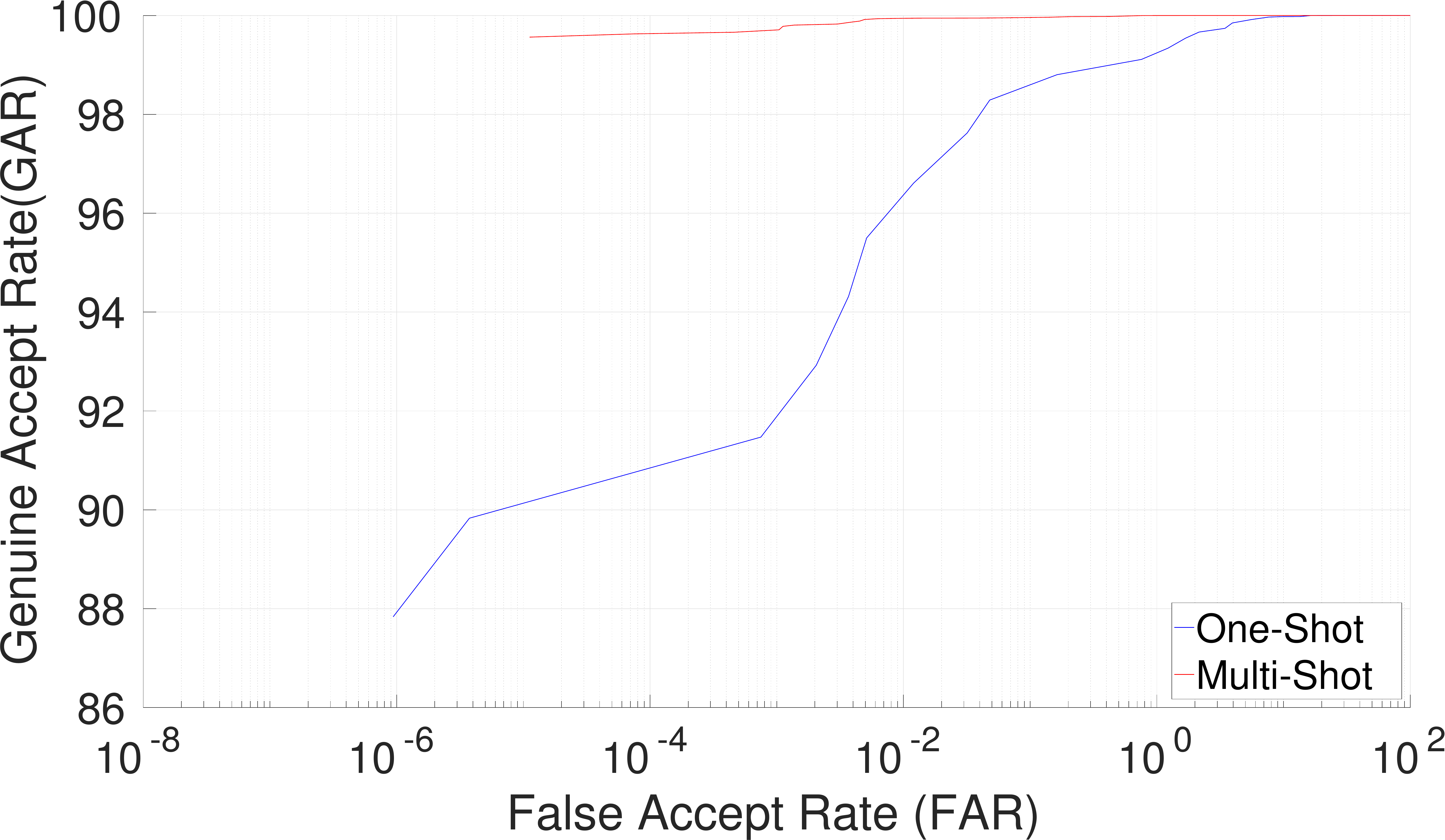}}
\subfigure[Yale]{\label{fig:b}\includegraphics[width=6.5cm,height=2.7cm]{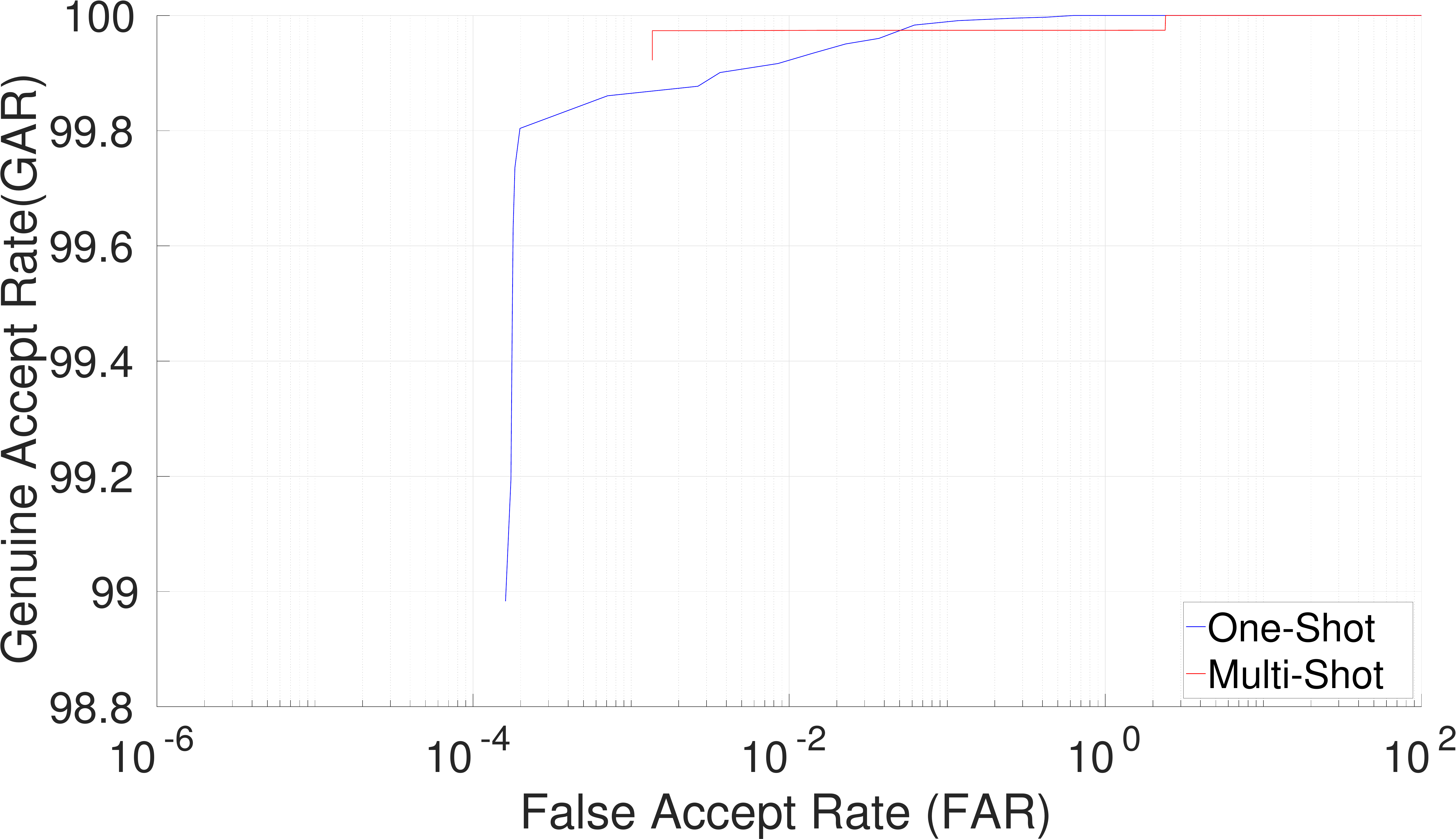}}
\subfigure[Multi-PIE]{\label{fig:c}\includegraphics[width=6.5cm,height=2.7cm]{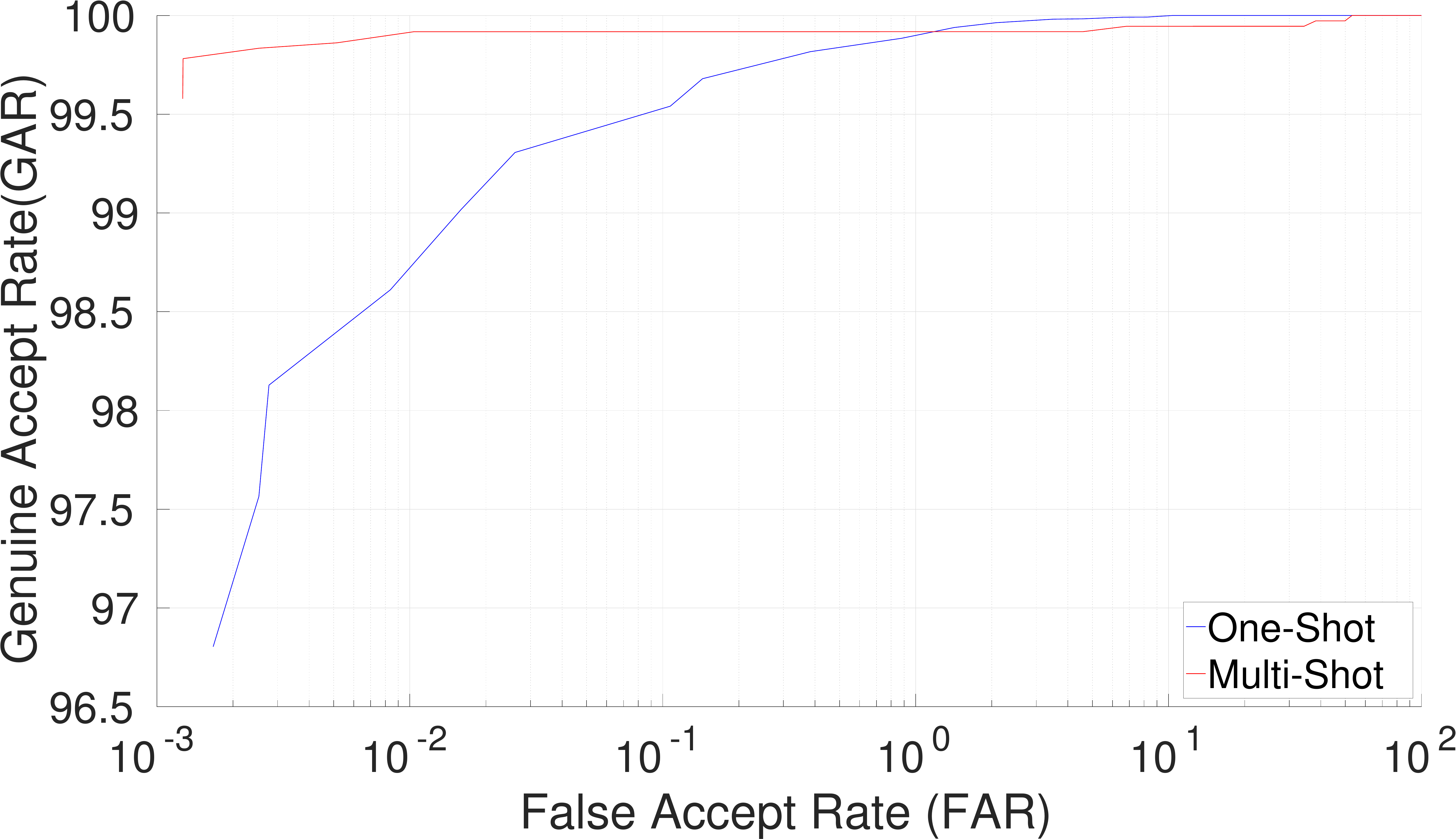}}
\subfigure[WVU]{\label{fig:c}\includegraphics[width=6.5cm,height=2.7cm]{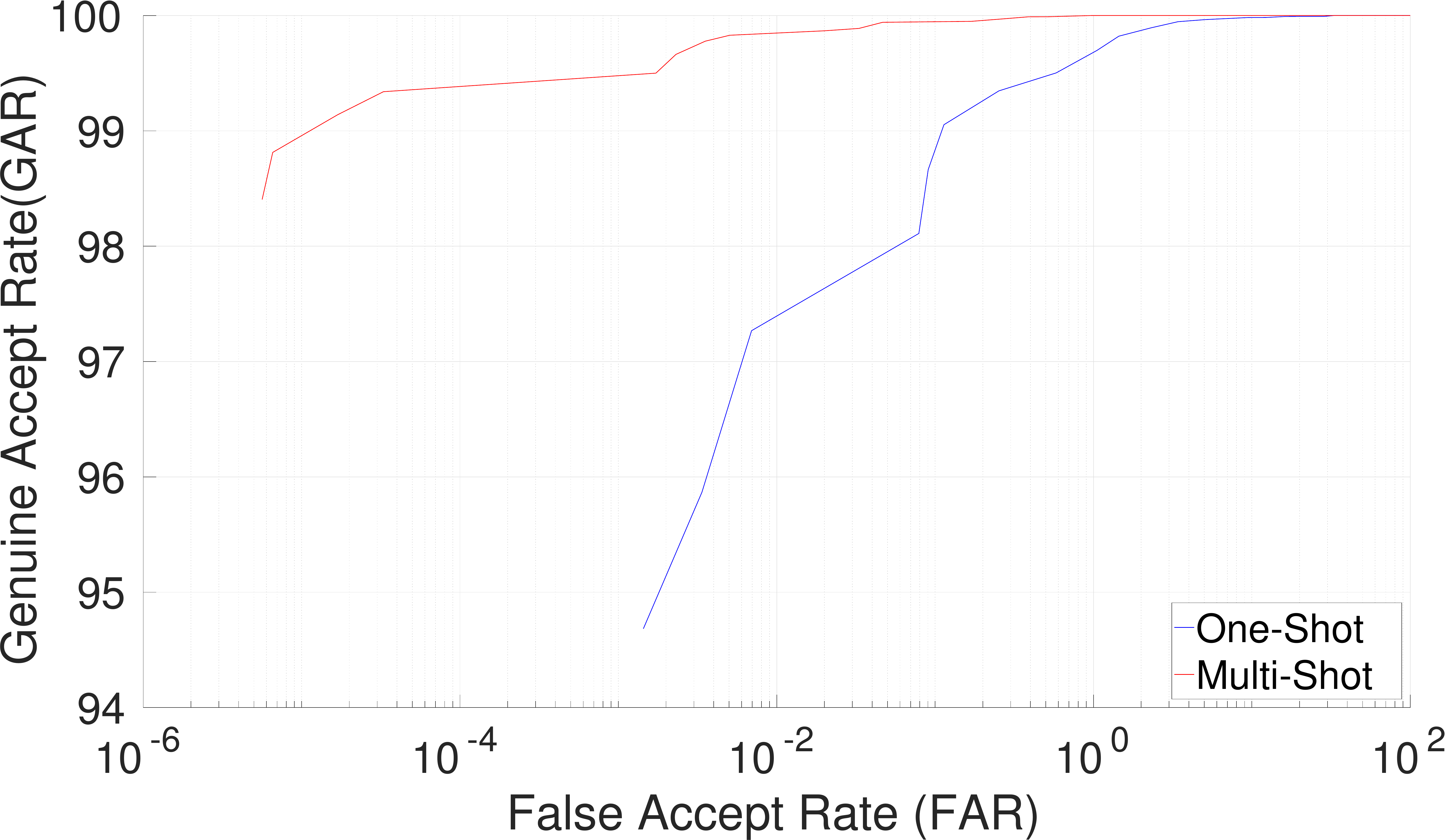}}
\vspace{-0.25cm}
\caption{ROC curves with One-Shot and Multi-Shot enrollment for different datasets for K=255.}
\label{fig:ROC_multi}
\vspace{-0.50cm}
\end{figure*}
\vspace{-0.30cm}


\subsection{Experimental Set Up and Results}
\vspace{-0.15cm}
We use the GAR at different FAR as the evaluation metric and we also report the equal error rate (EER). Since the train-test splits are randomly generated, we report the mean and standard deviation of the results for 5 random splits.

 As discussed in Sec. \ref{subsec:enroll}, authentication is based on binary accept/reject by comparing the probe template $\textbf{T}\textsubscript{p}$  with the enrolled template $\textbf{T}\textsubscript{e}$ .  However, this is not an ideal scenario for an experimental testing of biometric authentication system as we need a tunable metric to adjust the false accept rate (FAR) and false reject rate (FRR) of the system. For this reason, we use several augmented images (as described in Sec. \ref{subsec:augm})  of each image presented for authentication, and $\textbf{T}\textsubscript{p}$ is calculated for each augmented image, yielding a set of templates $\mathcal{T}$. Therefore, as given in \cite{pandey_deep_2016}, the final matching score can be defined by the number of templates $\textbf{T}\textsubscript{p}$ in $\mathcal{T}$ that match the stored template $\textbf{T}\textsubscript{e}$, scaled by the cardinality of $\mathcal{T}$. A threshold can then be applied to the matching score to achieve a desired value of FAR/FRR.

\begin{table}[t]
\centering
\scalebox{0.50}{\begin{tabular}{|c|c|c|c|c|}
 \hline
 \multicolumn{1}{|c}{\multirow{1}{*}{Database}} &\multicolumn{1}{|c}{\multirow{1}{*}{Enrollment Type}}
&\multicolumn{1}{|c}{\multirow{1}{*}{K}} &\multicolumn{1}{|c}{\multirow{1}{*}{$\mbox{GAR}@0.01\%\mbox{FAR}$}}
&\multicolumn{1}{|c|}{EER}\\ \hline
\multirow{6}{*}{PIE} & \multirow{2}{*}{Zero-Shot} &255& $83.6 \pm 2.1\%$ & $14.71 \pm 0.83\%$  \\ \cline{3-5}
& &1023&$82.8 \pm 1.83\%$ & $14.89 \pm 0.77\%$  \\ \cline{2-5}
&\multirow{2}{*}{One-Shot}&255&$96.2 \pm 0.98\%$ & $0.99 \pm 0.12\%$ \\ \cline{3-5}
& &1023&$96.0 \pm 1.12\%$ & $1.32 \pm 0.40\%$ \\ \cline{2-5}
&\multirow{2}{*}{Multi-Shot}&255&$99.9 \pm 0.06\%$ & $0.051 \pm 0.022\%$ \\ \cline{3-5}
& &1023&$99.0 \pm 0.88\%$ & $0.078 \pm 0.016\%$  \\ \hline
\multirow{6}{*}{Yale} & \multirow{2}{*}{Zero-Shot} &255& $87.4 \pm 1.46\%$ & $12.5 \pm 0.98\%$  \\ \cline{3-5}
& &1023&$85.1 \pm 1.98\% $ & $14.65 \pm 1.23\%$  \\ \cline{2-5}
&\multirow{2}{*}{One-Shot}&255&$99.9 \pm 0.03\%$ & $0.052 \pm 0.023\%$ \\ \cline{3-5}
& &1023&$99.1 \pm 0.18\%$ & $0.072 \pm 0.034\%$ \\ \cline{2-5}
&\multirow{2}{*}{Multi-Shot}&255&$99.98 \pm 0.005\%$ & $0.049 \pm 0.015\%$ \\ \cline{3-5}
& &1023&$99.8 \pm 0.09\%$ & $0.039 \pm 0.012\%$  \\ \hline
\multirow{6}{*}{Multi-PIE} & \multirow{2}{*}{Zero-Shot} &255& $81.4 \pm 2.32\%$ & $15.43 \pm 1.08\%$ \\ \cline{3-5}
& &1023&$81.2 \pm 2.18\%$ & $16.69 \pm 1.21\%$  \\ \cline{2-5}
&\multirow{2}{*}{One-Shot}&255&$98.7 \pm 0.99\%$ & $0.263 \pm 0.10\%$ \\ \cline{3-5}
& &1023&$97.4 \pm 0.94\%$ & $0.34 \pm 0.13\%$ \\ \cline{2-5}
&\multirow{2}{*}{Multi-Shot}&255&$99.8 \pm 0.17\%$ & $0.93 \pm 0.11\%$ \\ \cline{3-5}
& &1023&$98.5 \pm 0.43\%$ & $1.14 \pm 0.13\%$  \\ \hline
\multirow{6}{*}{WVU Multimodal} & \multirow{2}{*}{Zero-Shot} &255& $88.7 \pm 1.87\%$ & $10.32 \pm 0.83\%$ \\ \cline{3-5}
& &1023&$88.1 \pm 1.68\%$ & $11.32 \pm 0.92\%$  \\ \cline{2-5}
&\multirow{2}{*}{One-Shot}&255&$97.5 \pm 0.98\%$ & $0.42 \pm 0.14\%$ \\ \cline{3-5}
& &1023&$97.23 \pm 0.89\%$ & $0.51 \pm 0.11\%$ \\ \cline{2-5}
&\multirow{2}{*}{Multi-Shot}&255&$99.7 \pm 0.16\%$ & $0.11 \pm 0.02\%$ \\ \cline{3-5}
& &1023&$98.5 \pm 0.57\%$ & $0.48 \pm 0.10\%$  \\ \hline

\end{tabular}}
\vspace{-0.20cm}
\caption{Authentication results for different datasets.}
\captionsetup{width=.5\linewidth}
\label{table:ver_results}
\vspace{-0.75cm}
\end{table}

\begin{figure*}[h]
\centering     
\subfigure[Dictionary attack 1 ]{\label{fig:a}\includegraphics[width=5.4cm]{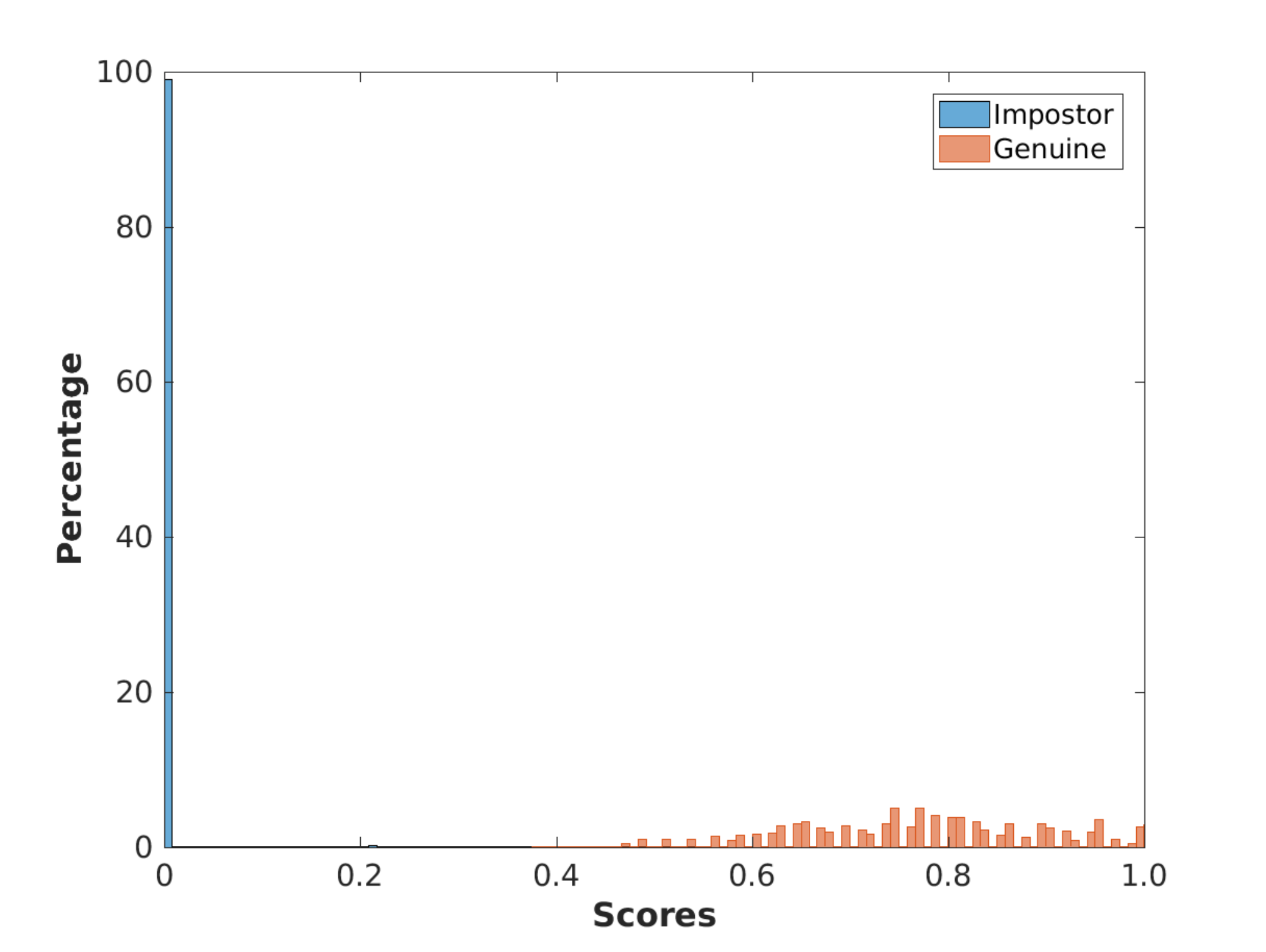}}
\subfigure[Dictionary attack 2]{\label{fig:b}\includegraphics[width=5.4cm]{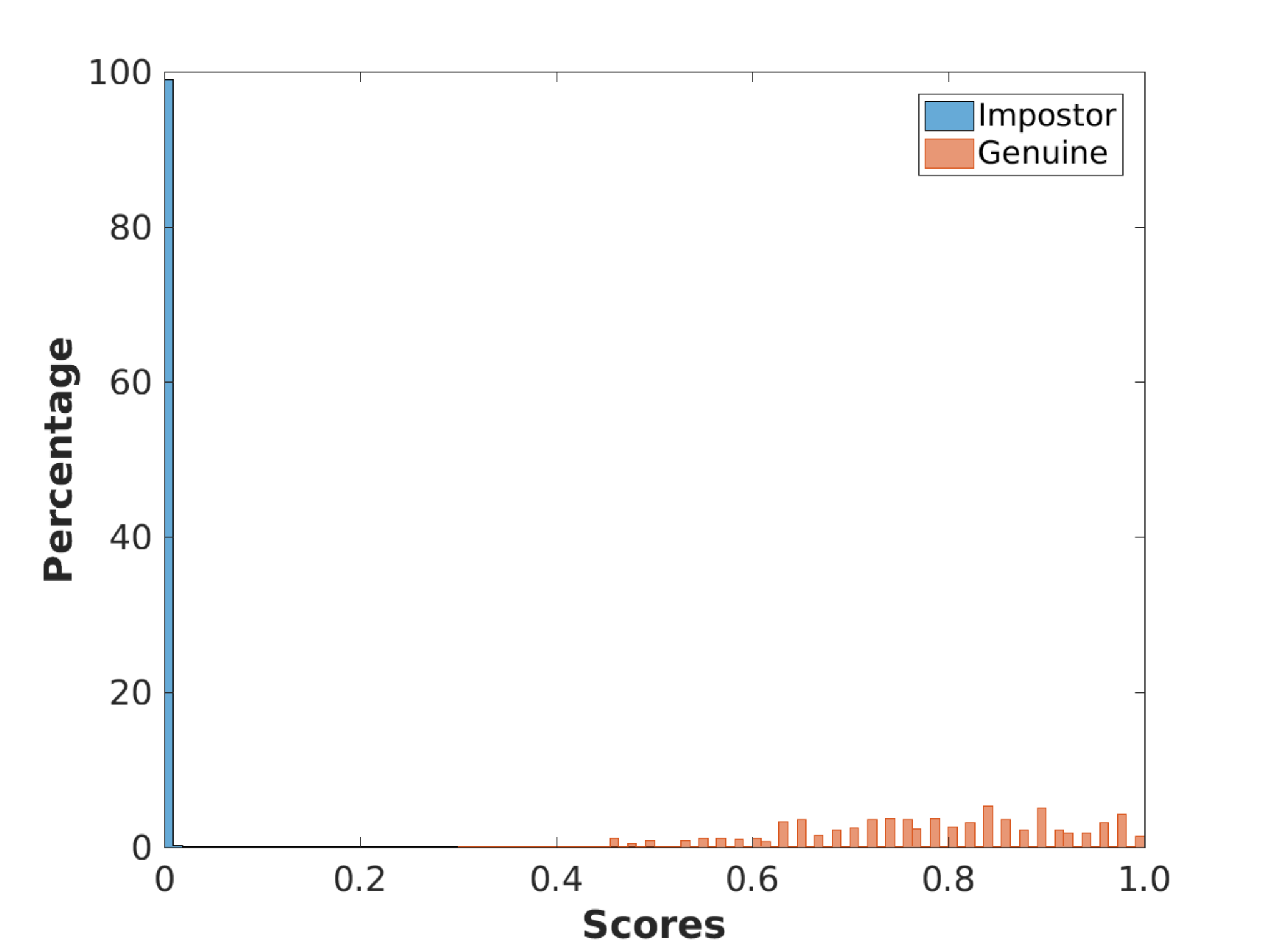}}
\subfigure[Dictionary attack 3]{\label{fig:c}\includegraphics[width=5.4cm]{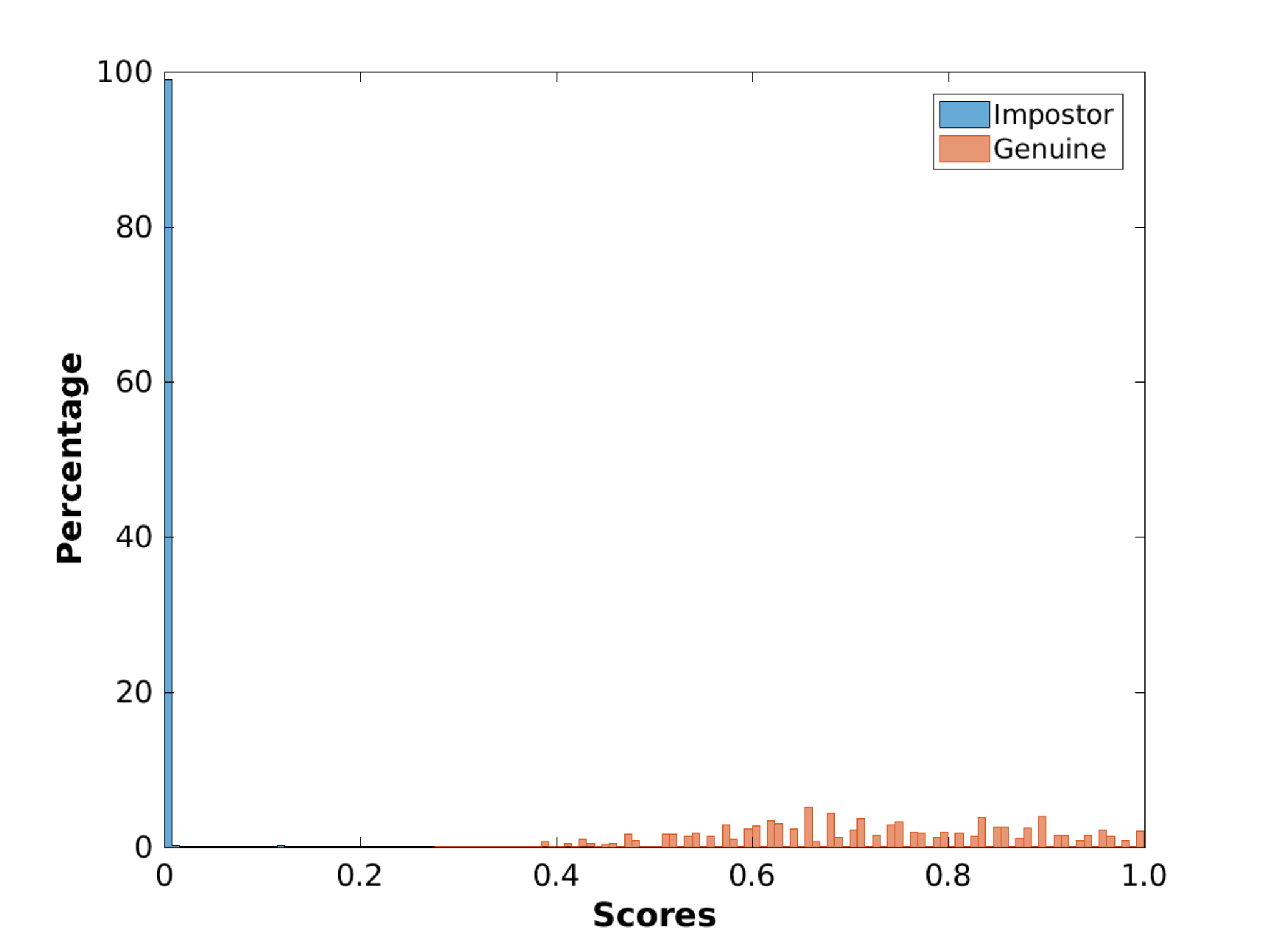}}
\vspace{-0.25cm}
\caption{Genuine and impostor distribution of for different dictionary attacks for K=255.}
\label{distr}
\vspace{-0.45cm}
\end{figure*} 

For our experiments, the mean and standard deviation of the EER, and the GAR at 0.01\% FAR for 5 different train-test splits, with zero-shot, one-shot, and multi-shot enrollment, using binary code dimensions $K=255,1023$ have been reported in Table \ref{table:ver_results}. With zero-shot enrollment, we achieve GARs up to $\approx 83\%$ on CMU-PIE, $\approx86\%$ on Extended Yale, $\approx81\%$ on Multi-PIE, and $\approx88\%$ on Multi-PIE  with up to $K=1023$ at the strict operating point of $0.01\%$ FAR. We also get high GARs in the range of $98-99\%$ for one shot and multi-shot on all the datasets. It can be observed from the Table \ref{table:ver_results} that the results are stable with respect to parameter $K$ as there is not drastic change in GAR or the EER as $K$ changes from 255 to 1023. Therefore, this makes the parameter $K$ totally selectable purely on the basis of the required template security. The verification performance using ROC curves is also shown in Fig. \ref{fig:ROC_multi} for all the datasets using one-shot and multi-shot.

\begin{table}[t]
\centering
\scalebox{0.50}{\begin{tabular}{|c|c|c|c|c|}
 \hline
 \multicolumn{1}{|c}{\multirow{1}{*}{Method}} &\multicolumn{1}{|c}{\multirow{1}{*}{Enrollment Type}}
&\multicolumn{1}{|c}{\multirow{1}{*}{K}} &\multicolumn{1}{|c}{\multirow{1}{*}{GAR@FAR}}
&\multicolumn{1}{|c|}{EER}\\ \hline
Hybrid Approach\cite{feng_2010_hybrid} &Multi-Shot&210&$90.61\%@1\%\mbox{FAR}$&$6.81\%$ \\ \hline
BDA\cite{feng_2012_bda} &Multi-Shot&76&$96.38\%@1\%\mbox{FAR}$&- \\ \hline
\multirow{2}{*}{MEB Encoding\cite{pandey_deep_2016}}&\multirow{2}{*}{Multi-Shot}&256&$93.22\%@0\%\mbox{\mbox{FAR}}$&$1.39\%$ \\ \cline{3-5}
& & 1024& $90.13\%@0\%\mbox{FAR}$&$1.14\%$ \\ \hline
\multirow{4}{*}{Deep CNN \cite{Jindal_2017_face}}&\multirow{2}{*}{Multi-Shot}&256&$97.35\%@0\%\mbox{FAR}$&$0.15\%$\\ \cline{3-5}
& & 1024& $96.53\%@0\%\mbox{FAR}$&$0.35\%$\\ \cline{2-5}
&\multirow{2}{*}{One-Shot}&256&$91.91\%@0.1\%\mbox{FAR}$&$4.00\%$ \\ \cline{3-5}
& & 1024& $91.34\%@0.1\%\mbox{FAR}$&$3.60\%$ \\ \hline
\multirow{6}{*}{Our Method}&\multirow{2}{*}{Multi-Shot}&\textbf{255}&$\textbf{99.9\%@0.01\%\mbox{FAR}}$&$\textbf{0.051\%}$\\ \cline{3-5}
& & \textbf{1023}& $\textbf{99.0\%@0.01\%\mbox{FAR}}$&$\textbf{0.078\%}$ \\ \cline{2-5}
&\multirow{2}{*}{One-Shot}&\textbf{255}&$\textbf{96.2\%@0.01\%\mbox{FAR}}$&$\textbf{0.99\%}$\\ \cline{3-5}
& & \textbf{1023}& $\textbf{96.0\%@0.01\%\mbox{FAR}}$&$\textbf{1.32\%}$ \\ \cline{2-5}
&\multirow{2}{*}{Zero-Shot}&\textbf{255}&$\textbf{83.6\%@0.01\%\mbox{FAR}}$&$\textbf{14.71\%}$\\ \cline{3-5}
& & \textbf{1023}& $\textbf{82.8\%@0.01\%\mbox{FAR}}$&$\textbf{14.89\%}$ \\ \hline

\end{tabular}}
\vspace{-0.20cm}
\caption{Performance comparison for PIE dataset}
\captionsetup{width=.5\linewidth}
\label{table:pie}
\vspace{-0.74cm}
\end{table}

A comparison of our results with other face template protection algorithms on PIE dataset is shown in Table \ref{table:pie}. Our proposed method values are all shown in bold. For security level, we compare our code dimensionality parameter $K$ to the equivalent parameter in the shown approaches. It can be noted that we get a better matching performance and a lower EER when compared to the other face template protection schemes for both one-shot and multi-shot enrollment. For one shot enrollment, we achieve $96.0\% \mbox{GAR} @ 0.01\% \mbox{FAR}$ for $K=1023$, which is $\approx 4.5\%$ improvement in matching performance compared to $91.34\% \mbox{GAR} @ 0.1\% \mbox{FAR}$ reported in \cite{Jindal_2017_face}. Even with zero-shot enrollment, we get good matching performance and a respectable EER.

\comment{In order to show the importance of NND, we have compared our complete architecture ``$DH+NND$" with two variations of our proposed system:1)Using the deep hashing component with no NND, denoted as ``$DH^{-}$". 2)DH component with an external conventional decoder, denoted as ``$DH+Decoder$". We tested these variations on WVU multimodal dataset for one -shot and multi-shot enrollment for 1023 bits. The EER results are shown in Table \ref{table:comp}. We can observe $DH+NND$ gives the best result followed by $DH+Decoder$ and lastly $DH^{-}$. We can see that using an external decoder improves the performance by reducing the EER by about $5\%$ and using the NND further reduces EER by $1.1\%$. The advantage with NND over external conventional decoder is that it provides an opportunity to jointly learn and optimize with respect to biometric datasets, which  are  not  necessarily  characterized  simply  by  Gaussian noise as is assumed by a conventional decoder}.   

\begin{table}[t]
\centering
\scalebox{0.75}{\begin{tabular}{|c|c|c|c|}
 \hline
 \multicolumn{1}{|c}{\multirow{1}{*}{Enrollment Type}}
&\multicolumn{3}{|c|}{\multirow{1}{*}{Method}}\\ \cline{2-4}
&$DH^{-}$&$DH+Decoder$&$DH+NND \text{(Our)}$\\ \hline
One-Shot&$ 8.2\%$&$1.83\%$&$0.51\%$ \\ \hline
Multi-Shot&$ 6.61\%$&$1.54\%$&$0.48\%$\\ \hline
\end{tabular}}
\vspace{-0.20cm}
\caption{EER comparison with other variations for WVU multimodal  dataset for 1023 bits}
\captionsetup{width=.5\linewidth}
\label{table:comp}
\vspace{-0.50cm}
\end{table}

\vspace{-0.25cm}

\section{Security Analysis}
\vspace{-0.15cm}
This security analysis is based on a stolen template scenario as reported in \cite{pandey_deep_2016}. Given the template, the attacker's goal is to extract biometric information of the user. However, the template is generated using a cryptographic hash function SHA-3 512 of the binary codes generated by the the deep CNN (DH + NND). Morover, the binary codes generated are neither stored nor provided to the user. Since the SHA-3 512 is a one-way transformation, the attacker will not be able to extract any information about the binary codes from the protected template. The only way the attacker can get the codes is by a brute force attack, where the attacker would need to try all possible values of the code, hash each one and compare them to the template.

There are two scenarios that need to be explored in the case of brute force attack. The first scenario is when the attacker has no access to the CNN parameters and the second scenario is when the attacker has access to the CNN parameters. In the scenario where the attacker has no access to the CNN parameters, the search space for the brute force attack would be $2^K$, i.e., the number of binary codes. In this scenario, it is very important that the final binary code should posses high entropy. The objective function required to train the DH component also includes a loss function for maximizing the entropy, which is shown in (\ref{eq:4}). Therefore, the high entropy requirement is captured in the training of the DH component. Additionally, to make the search space larger for a brute force attack, we use a final binary code with a minimum dimensionality of $K=255$. The brute force attack in this scenario would be computationally infeasible because even with $K=255$, the search space would be the order of $2^{255}$ or larger. 

Now, let's analyze the scenario where the attacker has access to both the stolen face protected template and also the CNN parameters. In this scenario, the attacker would try to generate attacks in the input domain and exploit the FAR of the system. To exploit the system FAR, the attacker would try a dictionary attack using large set of faces. In the proposed method, it is not straightforward to analyze the reduction in the search space due to the knowledge of the CNN parameters. However, measuring the minimal FAR of the proposed method is a good indicator of the template security.  To evaluate the template security, we have used the genuine and impostor distribution to measure the false accepts, where all other users other than genuine are considered as impostors. The genuine and impostor distributions for three types of dictionary attacks are shown in Fig. \ref{distr}. The three attacks are:(1) CMU-PIE as the genuine database and Ext Yale as the attacker database. (2) Ext Yale as a genuine database and frontal images of Multi-PIE as attacker database (3) Multi-PIE as the genuine database and Ext Yale as the attacker database. It can be seen that the impostor scores are always zero and genuines tend to one, indicating there are no false accepts in this scenario and the proposed method does not easily accept external faces for enrolled faces even if they are preprocessed under same conditions. 
\vspace{-0.35cm}
\section{Conclusion}
\vspace{-0.15cm}
In this work, we present an algorithm that uses a combination of deep hashing and neural network based error correction to be implemented for face template protection. The novelty of this algorithm is that it can even be used for zero-shot enrollment, where the subject has not been seen during training of the deep CNN and still can be enrolled. Additionally, we show a matching performance improvement of $\approx 4.5\%$ for one-shot enrollment and $\approx 3\%$ for multi-shot enrollment when compared to related work, while providing high template security.    
\vspace{-0.25cm}
{\small
\bibliographystyle{ieee}
\bibliography{egbib}
}

\end{document}